%% file: arxiv.tex
\documentclass[10pt,twocolumn,letterpaper]{article}
\usepackage{iccv}
																									 				  
\usepackage{times}
\usepackage{epsfig}
\usepackage{graphicx}
\usepackage{amsmath}
\usepackage{amssymb}
\usepackage{makecell}
\usepackage{caption}
\usepackage{wrapfig}
\usepackage{booktabs}

\usepackage{multirow}
\usepackage{mathptmx}

\newcommand{\ignore}[1]{{}}

\usepackage[pagebackref=true,breaklinks=true,letterpaper=true,colorlinks,bookmarks=false]{hyperref}

\usepackage[pagebackref=true,breaklinks=true,letterpaper=true,colorlinks,bookmarks=false]{hyperref}

\iccvfinalcopy 
\def\iccvPaperID{5959} 

\usepackage{multirow}

\usepackage{caption}
\usepackage{subcaption}

\usepackage[T1]{fontenc}
\usepackage[utf8]{inputenc}
\usepackage{authblk}

\begin{document}

\title{EgoRenderer: Rendering Human Avatars from Egocentric Camera Images}

%
%
%



\author{Tao Hu$^{1}\thanks{Work partly conducted during TH's internship at MPI-INF}$, ~ Kripasindhu Sarkar$^2$, ~Lingjie Liu$^2$, ~Matthias Zwicker$^1$, ~Christian Theobalt$^2$ \\
	\vspace{-0.08in}
	$^1$Department of Computer Science, University of Maryland, College Park \\ 
	$^2$Max Plank Institute for Informatics, Saarland Informatics Campus \\
}


\maketitle
\thispagestyle{empty}

\begin{abstract}
	
	We present EgoRenderer, a system for rendering full-body neural avatars of a person captured by a wearable, egocentric fisheye camera that is mounted on a cap or a VR headset. 
	Our system renders photorealistic novel views of the actor and her motion from arbitrary virtual camera locations. Rendering full-body avatars from such egocentric images come with unique challenges due to the top-down view and large distortions. 
	We tackle these challenges by decomposing the rendering process into several steps, including texture synthesis, pose construction, and neural image translation. 
	For texture synthesis, we propose Ego-DPNet, a neural network that infers dense correspondences between the input fisheye images and an underlying parametric body model, and to extract textures from egocentric inputs. In addition, to encode dynamic appearances, our approach also learns an implicit texture stack that captures detailed appearance variation across poses and viewpoints. For correct pose generation, we first estimate body pose from the egocentric view using a parametric model. We then synthesize an external free-viewpoint pose image by projecting the parametric model to the user-specified target viewpoint. We next combine the target pose image and the textures into a combined feature image, which is transformed into the output color image using a neural image translation network. Experimental evaluations show that EgoRenderer is capable of generating realistic free-viewpoint avatars of a person wearing an egocentric camera. Comparisons to several baselines demonstrate the advantages of our approach. Project page: \url{https://vcai.mpi-inf.mpg.de/projects/EgoRenderer/}.
\end{abstract}

\begin{figure}
	\begin{center}
		\includegraphics[width=\linewidth]{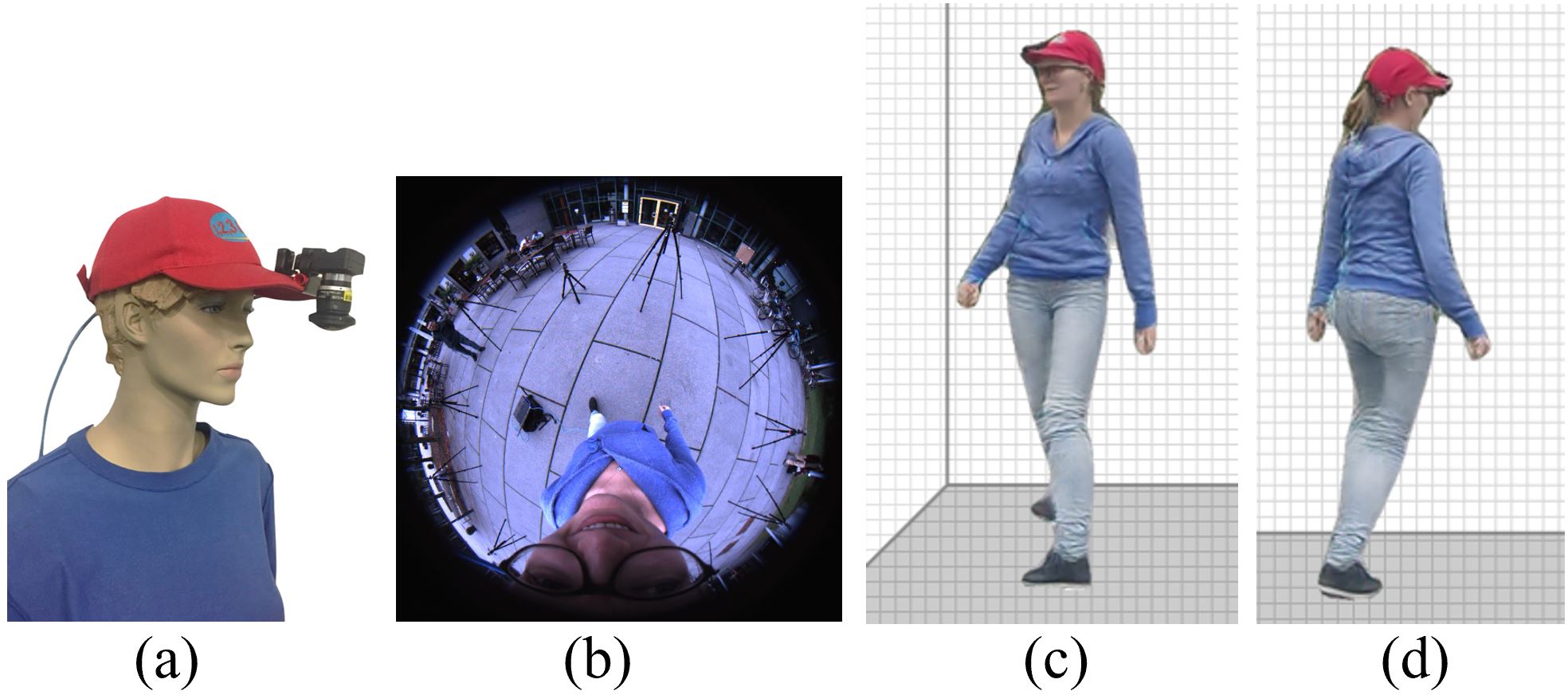}
	\end{center}
	\vspace{-0.16in}
	\caption{Based on a wearable fisheye camera setup (a), we propose EgoRenderer, which is trained for a single person and can produce full-body avatars of the person from new viewpoints and in new poses (c)(d) by taking as input an  egocentric image (b) captured by the fisheye camera.}
	\label{fig:teaser}
	\vspace{-0.15in}
\end{figure}

\section{Introduction}

\input{intro_new.tex}

\input{related_work.tex}

\section{Approach}

Our EgoRenderer system is built on a single cap-mounted fisheye camera (similar to \cite{mo2cap2}) that captures egocentric images of persons at run time. Given an egocentric image $I_e$ of a person, we render a full body avatar of the person from a user-defined viewpoint, which we call external free-viewpoint space in this paper. Our system decomposes the rendering process into texture synthesis, pose construction, and neural image translation, as shown in Figure \ref{fig:overview}. First, for texture synthesis, we build a global texture stack $T_g$ that combines explicit textures $T_e$ from egocentric images and an implicit texture stack $T_m$, which is learned in a training phase, as the texture representations of the person. Our system represents body pose and shape as a parametric mesh~\cite{smpl}. For pose construction, given a user-defined target viewpoint, we synthesize a target pose image by projecting the parametric model from egocentric camera space to the target viewpoint. The third step renders the 3D model with the global texture stack and generates implicit feature images $R_{e \rightarrow t}$ of the person, which are then transformed to the final rendered color image $R_{e \rightarrow t}$ by a neural image-translation network $RenderNet$ in the fourth step. 



\subsection{Input and Output}
Our input is an egocentric image $I_e$ of a person and a target viewpoint $v$, and the output is a photorealistic image $I_{e \rightarrow t}$ of the person from the target viewpoint. In our training stage, we take pairs of images ($I_e$, $I_t$) of the same person as input. In this paper, we call $I_e$ images from egocentric space, and $I_t$ images from external free-viewpoint space.

\subsection{Texture Synthesis}

\noindent \textbf{Extracting Partial UV Texture Maps from Egocentric Space.} The pixels of the input egocentric images are transformed to UV texture space through dense correspondences between the input images and an underlying SMPL~\cite{smpl} model. DensePose \cite{densepose} (pre-trained on the COCO-DensePose dataset), can predict 24 body segments and their part specific UV coordinates on images captured with regular cameras and mostly from chest high viewpoints. However, DensePose \cite{densepose} fails on the egocentric images in our setup, as shown in Figure \ref{fig:comp_dp}. 

Capturing a large amount of annotated DensePose data for egocentric data is a mammoth task that would require large amounts of human labor. To solve this problem, we rendered a large, synthetic egocentric fisheye training dataset that enables training of a deep neural network to predict dense correspondences on our hardware setup. Our dataset contains 178,800 synthetic images with ground truth UV coordinate maps, and features various poses, body appearances and backgrounds, as shown in Figure \ref{fig:synthetic}. With this synthetic dataset, we train an Ego-DPNet ($f$) network to predict the DensePose $P_e$, that is,
$P_e=f(I_e)$. More details of the dataset can be found in the supplementary material.

Ego-DPNet was built on the DensePose-RCNN architecture, and we trained it in multi-stage using transfer learning. Different from the original DensePose-RCNN, which takes sparse annotated ground truth points and relies on a inpainting teacher network to interpolate dense ground truth, our Ego-DPNet directly takes synthetic images as dense ground truth without the teacher network. In addition, another difference is that the input image size of Ego-DPNet is fixed in training and testing due to our specific fisheye camera setup. The performance of Ego-DPNet is shown in Figure \ref{fig:comp_dp}. With the DensePose prediction $P_e$ and the input image $I_e$, we extract the partial UV texture map $T_e$ by an \textit{UV Texture Extraction} module $u$, that is, $T_e = u(I_e, P_e)$.  

\begin{figure}
	\begin{center}
		\includegraphics[width=0.8\linewidth]{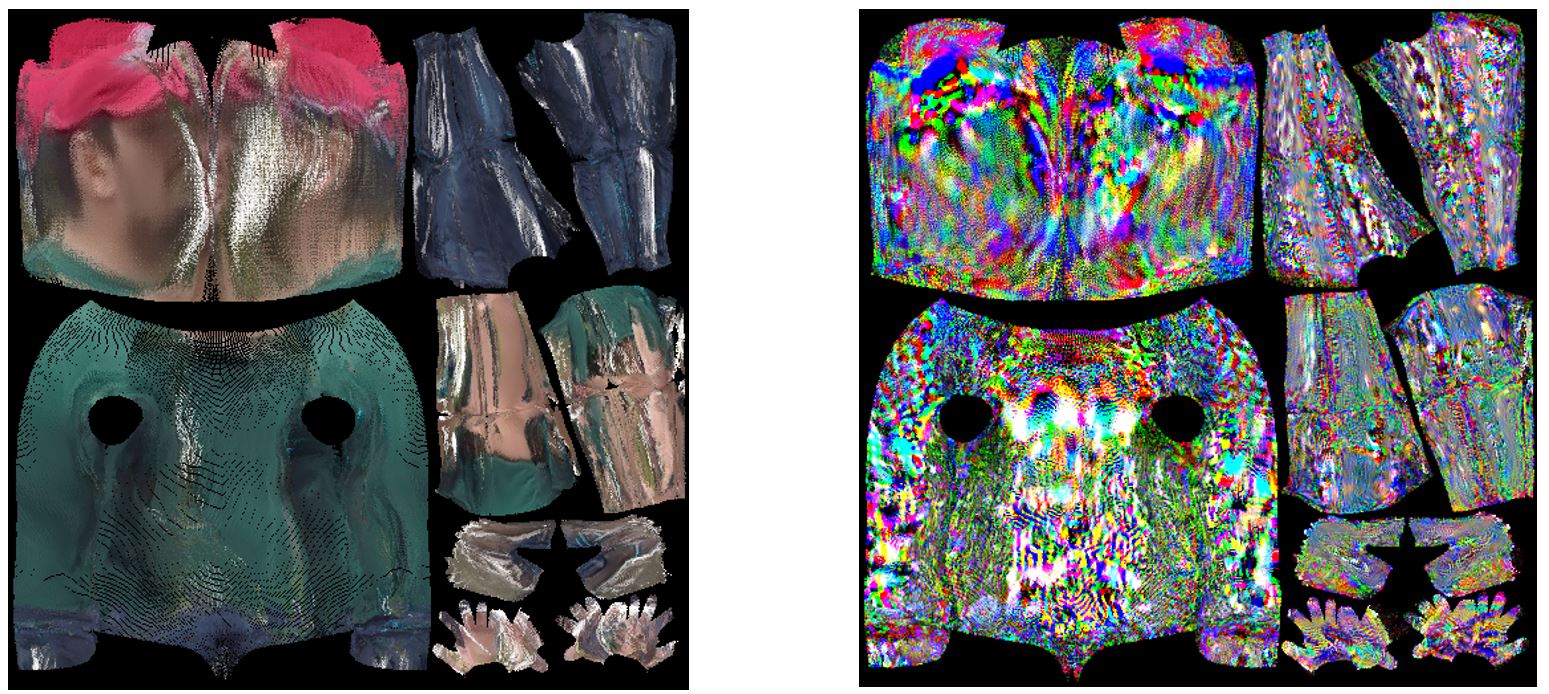}
	\end{center}
	\vspace{-0.17in}
	\caption{The impact of learning on the implicit texture stack $T_m$. Left: initial state; right: final state.}
	\label{fig:implicit_tm}
	\vspace{-0.15in}
\end{figure}

\noindent \textbf{Learning Implicit Textures in Training.}
Besides textures from egocentric space, we also learn a $d$-dimensional implicit texture stack ($T_m$) from training images to capture detailed appearance variation across poses and viewpoints. Note that we initialize $T_m$ with an explicit texture map that is the average texture of the training images. The initial and final state of $T_m$ during training is shown in Figure \ref{fig:implicit_tm}. We provide our experiments with 3 channels in $T_m$.

\noindent \textbf{Global Texture Stack.} We concatenate explicit textures $T_e$ and implicit textures $T_m$ into the \textit{global texture stack} $T_g = [T_e, T_m]$ to represent dynamic textures of the person. In our paper, $T_g$ has 6 channels, 3 from $T_e$ and 3 from $T_m$.

\subsection{Pose Construction}
\label{sec:pose}
Given an egocentric image $I_e$ and a target user-defined viewpoint $v$, the second step is to synthesize the target pose of the person under the viewpoint, which is represented by a DensePose image $P_t$ in Figure \ref{fig:overview}. First, a pose estimation module (Mo2Cap2 \cite{mo2cap2} in our experiments) is used to extract the 3D joint pose of the person, which is used to drive a 3D SMPL model by \textit{Inverse Kinematics}. The target pose is then rendered by projecting the 3D model to viewpoint $v$.

\subsection{Intermediate Feature Image Rendering}

Using the global texture stack $T_g = [T_e, T_m]$ and the target pose image $P_t$, we use a \textit{Feature Rendering} operation $r$ to produce a 6-dimensional feature image $R_{e \rightarrow t}$, that is, $R_{e \rightarrow t} = r(T_e, T_m, P_t)$. The operation $r$ is implemented by differentiable, bilinear sampling in our experiments. 

\subsection{Neural Image Translation}

In the final step, the feature image $R_{e \rightarrow t}$ is translated to a realistic image $I_{e \rightarrow t}$ using a translation network $g$, which we call RenderNet, $I_{e \rightarrow t} = g(R_{e \rightarrow t})$. RenderNet is built on the Pix2PixHD \cite{pix2pixhd} architecture. The discriminator for adversarial training of RenderNet
also uses the multiscale design of Pix2PixHD, and we use a three scale discriminator network for the adversarial training.

\subsection{Training Details and Loss Functions}
We train the EgoRenderer in two stages. We first train the Ego-DPNet on our synthetic dataset, and then train RenderNet on real data. Note that RenderNet is person spcific.

\noindent \textbf{Ego-DPNet}. For better generalization on real world
imagery, we train Ego-DPNet in multiple stages using transfer learning. It was first pre-trained on the COCO-DensePose dataset \cite{densepose} to learn good low-level features from real images with usual camera optics. Then we fine tune it on our synthetic dataset to predict DensePose on egocentric images. 

\noindent \textbf{RenderNet}. In training our system takes pairs of egocentric inputs and ground truth images ($I_e$, $I_t$) of the same person as input. The output of RenderNet can be expressed as
$$
\vspace{-0.08in}
I_{e \rightarrow t} = g \circ r (u(I_e, f\left(I_{e}\right)), T_m, P_t), 
$$

where all operations $g, r, f, u$ are differentiable. For speed, we pre-compute Ego-Pose $P_e=f\left(I_{e}\right)$ and $P_t$, and directly read them as input in training. We optimize the parameters of RenderNet $g$ and the implicit texture stack $T_m$. 

\noindent \textbf{Loss Functions.} We apply the combination of the following loss functions to train RenderNet:

$\bullet$ \textbf{Perceptual Loss}. We use a perceptual loss based on the VGG Network \cite{Perceptual_Losses}, which measures the difference between the activations on different layers of the pretrained VGG network \cite{vgg} applied on the generated image $I_{e \rightarrow t}$ and ground truth target image $I_t$,
$$
\vspace{-0.08in}
L_{p}=\sum \frac{1}{N^{j}}\left|p^{j}\left(I_{e \rightarrow t}\right)-p^{j}\left(I_{t}\right)\right|,
$$

where $p^j$ is the activation and $N^j$ the number of elements of the $j$-th layer in the pretrained VGG network.

$\bullet$ \textbf{Adversarial Loss}. We use a multiscale discriminator $D$ of Pix2PixHD \cite{pix2pixhd} to leverage an adversarial loss $L_{adv}$ in our system. $D$ is conditioned on both the generated image and rendered feature image.

$\bullet$ \textbf{Face Identity Loss}. We use a pre-trained network to ensure that RenderNet and the implicit texture stack preserve the face identity on the cropped face of the generated and ground truth image,
$$
\vspace{-0.08in}	
L_{{face}}=\mid N_{ {face}}\left(I_{e \rightarrow t}\right)-N_{ {face}}\left(I_{t}\right)|,
$$

where, $N_{face}$ is the pretrained SphereFaceNet \cite{Liu2017SphereFaceDH}.

\noindent The final loss is then
$$
\vspace{-0.08in}
L_{G}=\lambda_{p} L_{p} + \lambda_{face} L_{face} + \lambda_{GAN} L_{adv}.
$$

The networks are trained using the Adam optimizer \cite{adam} with an initial learning rate of $2\times 10^{-4}$, $\beta_1 = 0.5$. The loss weights are set empirically to $\lambda_{GAN} = 1$, $\lambda_{p} = 10$, $\lambda_{face} = 5$. Note that the initial learning rate of $T_m$ is $2\times 10^{-3}$, 10 times of the learning rate of RenderNet.

\section{Experiments}


\begin{figure}
	\begin{center}
		\includegraphics[width=0.9\linewidth]{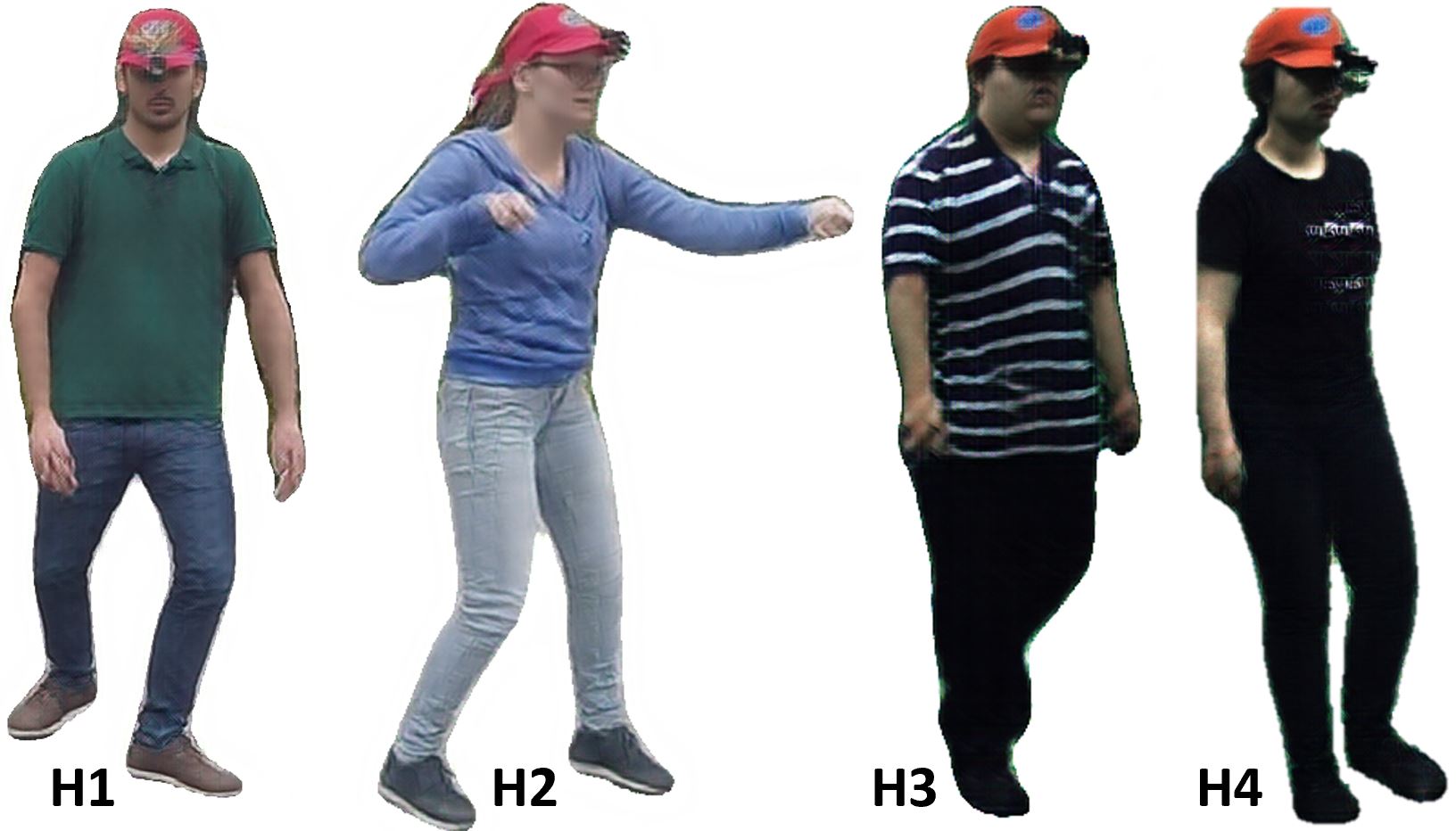}
	\end{center}
	\vspace{-0.2in}
	\caption{Renderings produced by our methods (for all subjects H1, H2, H3, H4 in our study). All renderings are produced from new viewpoints and poses unseen in training.}
	\label{fig:all_avatar}
	\vspace{-0.18in}
\end{figure}


%

\begin{figure*}	
	\begin{center}
		\includegraphics[width=\linewidth]{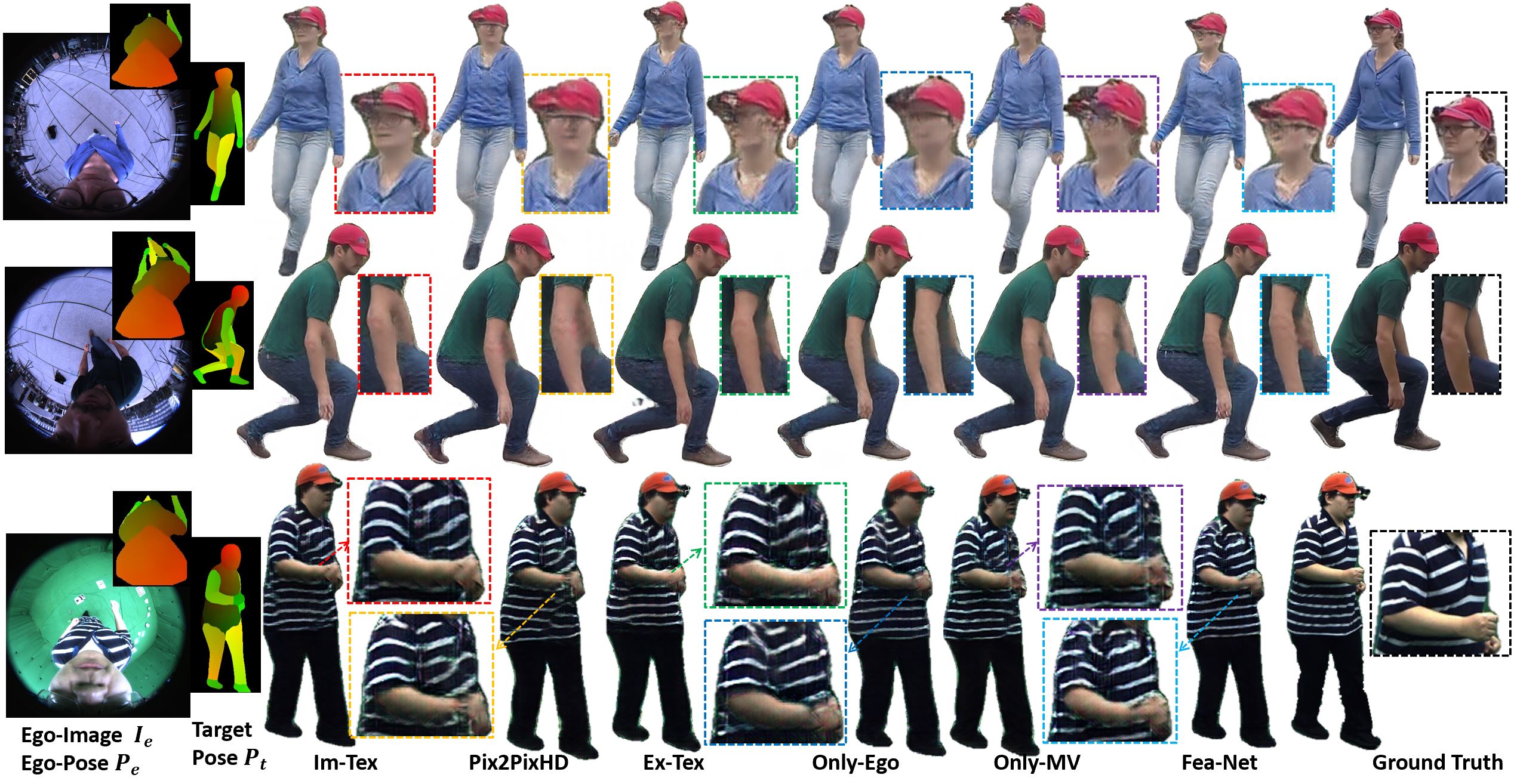}
	\end{center}
	\vspace{-0.18in}
	\caption{Comparisons of the six methods. All avatars are generated from novel viewpoints and poses unseen in training.}
	\label{fig:comp_method}
\end{figure*}



\noindent \textbf{Datasets.} Since there are no public datasets suitable for our project, we captured 4 datasets by ourselves, and the rendered human avatars in our study are shown in Figure \ref{fig:all_avatar}. We refer to them as H1, H2, H3 and H4. H1 and H2 were captured in outdoor scenes, while H3 and H4 were captured in a studio. We have 11 multi-view cameras for the H1 and H2 datasets, and 8 for H3 and H4, and use the training/test split of 80\%/20\% to train and evaluate our methods.

\subsection{Baselines}
\label{sec:baselines}

We denote our method as Im-Tex, and compare Im-Tex with 5 other systems, Pix2PixHD \cite{pix2pixhd}, Fea-Net \cite{feanet}, and three variants, Ex-Tex, Only-Ego, and Only-MV. All these baseline methods have similar architectures as ours, and we provide them with the same input and loss functions in training, so all these methods are directly comparable. 1) \textbf{Pix2PixHD.} We used Pix2PixHD's code with minimal modifications that lead to better performance. Pix2PixHD directly translates a target pose $P_t$ to a generated image $I_{e \rightarrow t}$. 2) \textbf{Fea-Net.} We refer to the human re-rendering method \cite{feanet} as Fea-Net (short for FeatureNet), which is the state of the art in human re-enactment and novel view rendering when this project was developed. Different from our method, which maintains a global implicit texture stack, Fea-Net uses a network to extract 16-dimensional feature maps from egocentric images that serve as implicit textures. Though Fea-Net was originally trained on multiple identities to do garment and motion transfer, it can also be directly applied on our person specific task. 3)  \textbf{Ex-Tex.} As a variant of our implicit texture method, Ex-Tex works with an explicit and static texture stack. Compared with Im-Tex, the texture stack ($T_m$) in Figure \ref{fig:overview} is not updated during training. Hence we call it an explicit texture stack as shown in Figure~\ref{fig:implicit_tm}-left. 4) \textbf{Only-Ego.}
We also consider a variant that does not have the texture stack $T_m$, which means $T_g=T_e$, and in feature rendering, we only sample textures from egocentric images. We call this method Only-Ego. 5) \textbf{Only-MV.}
In contrast to Only-Ego, another variant is Only-MV, which does not take textures from egocentric images, but only from training images $I_t$. In this case, $T_g=T_m$. Only-MV can be seen as an extension of Deferred Neural Rendering \cite{Thies2019}.

\begin{table*}[ht]
	\begin{tabular}{|c|c|c|c|c|c|c|c|c|c|c|c|}
		\hline
		\multicolumn{1}{|c|}{\multirow{2}{*}{Methods}} & \multicolumn{2}{c|}{H1} & \multicolumn{2}{c|}{H2} & \multicolumn{2}{c|}{H3} & \multicolumn{2}{c|}{H4} & \multicolumn{3}{c|}{Mean relative improvements(\%)} \\ \cline{2-12} 
		\multicolumn{1}{|c|}{}                         & SSIM$\uparrow$  & LPIPS$\downarrow$  & SSIM  & LPIPS  & SSIM & LPIPS  & SSIM  & LPIPS  & $RI_{SSIM}\uparrow$ & $RI_{LPIPS}\uparrow$ & $RI_{PSNR}\uparrow$ \\ \hline
		Im-Tex                                         & 7.529  & \textbf{1.623}  & \textbf{6.840}  & \textbf{1.617}  & \textbf{6.469}  & \textbf{1.569}  & 7.535  & \textbf{1.571} & \textbf{1.268} & \textbf{7.562} & .748\\ \hline
		Pix2PixHD                                      & 7.395  & 1.713  & 6.798  & 1.713  & 6.342  & 1.640  & 7.428  & 1.615 & - & 2.427 & .404 \\ \hline
		Ex-Tex                                         & 7.431  & 1.691  & 6.778  & 1.683  & 6.326  & 1.676  & 7.435  & 1.629 & 0.089  & 3.630 & .555 \\ \hline
		Only-Ego                                       & 7.437  & 1.769  & 6.782  & 1.704  & 6.361  & 1.660  & \textbf{7.586}  & 1.578 & 0.240 & - & .467 \\ \hline
		Only-MV                              & \textbf{7.543}  & 1.738  & 6.828  & 1.626  & 6.360  & 1.585  & 7.505  & 1.587 & 1.190 & 4.263 & - \\ \hline
		Fea-Net                                        & 7.444  & 1.695  & 6.823  & 1.687  & 6.350  & 1.630  & 7.449  & 1.616 & 0.487 & 3.600 & \textbf{.912} \\ \hline
		\end{tabular}
		\vspace{-0.1in}
		\caption{Quantitative results (multiplied by 10) of single-video training on different datasets, and relative improvements.}
		\vspace{-0.16in}
		\label{tab:quant_single}
		\end{table*}

\subsection{Evaluations on Test Datasets}

We consider two regimes: training on single- or multi-camera video sequences. The evaluation is done on the hold-out cameras and hold-out parts of the sequence. That is, there is no overlap between the training and test sets in terms of the camera or body poses. For fair comparisons among different methods, we use DensePose results computed on the target frame as input in training and evaluation like other human rendering papers \cite{texava,everybody}.

\begin{figure*}	
	\begin{center}
		\includegraphics[width=0.95\linewidth]{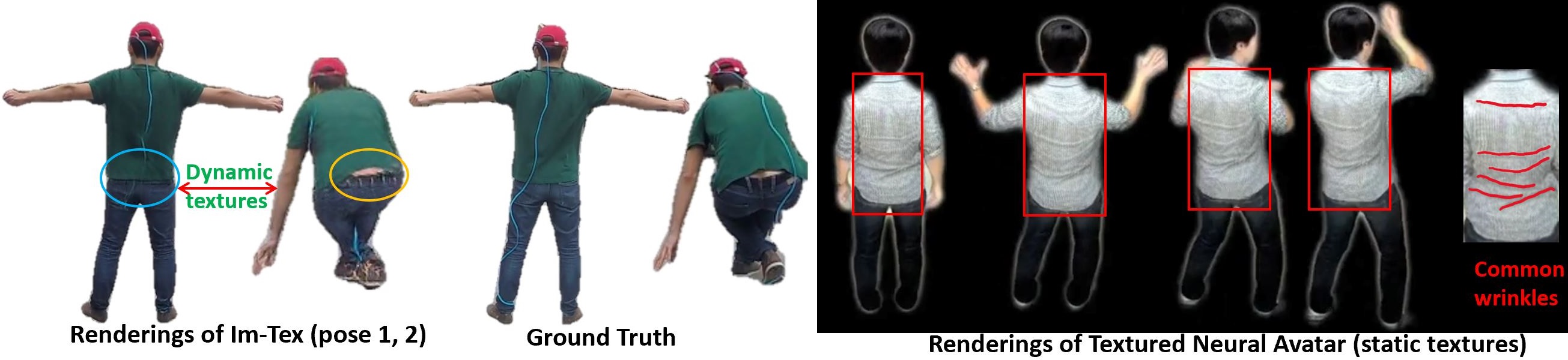}
	\end{center}
	\vspace{-0.22in}
	\caption{Comparisons to Textured Neural Avatars \cite{texava}. Our method (left) produces realistic time- and pose-dependent appearance detail, such as pose-dependent wrinkles in clothing, or the shifting of the shirt. In contrast, Textured Neural Avatars cannot produce the same level of realism, since a static texture with fixed clothing wrinkles is warped into new poses.}
	\label{fig:comp_tna}
	\vspace{-0.16in}
\end{figure*}

\noindent \textbf{Metrics.} The following three metrics are used for comparison, the Structural Similarity Index (SSIM) \cite{ssim}, and the Learned Perceptual Image Patch Similarity (LPIPS) \cite{lpip}, and peak signal-to-noise ratio (PSNR). The recent LPIPS claims to capture human judgment better than existing hand designed metrics, whereas PSNR or SSIM is often inaccurate in capturing the visual perception and would differ starkly with slight changes in texture/wrinkles \cite{lpip}.



\noindent \textbf{Single video experiments.} 
We first evaluate our system in a single video case. We use single-camera videos from one of the cameras in our rig, and evaluate the 6 methods on the hold-out cameras and hold-out parts of the sequence. The qualitative results are shown in Figure \ref{fig:comp_method}. It can be observed that our results show better realism and preserve more details, and the quantitative results are provided in Table \ref{tab:quant_single}. We also calculate the mean relative improvement ($RI$) of each approach ($x$) over the worst one ($y$, indicated by -) on metric $m$ for all datasets: $RI_{m}(x, y) = |m(y)-m(x)|/m(y)$, where $m(x)$ is the result of $x$ on metric $m$. $RI$ is shown in percent. Our Im-Tex outperforms the others on 8 out of 11 metrics, and we achieve the best LPIPS scores on all the datasets. In general, our system is capable of learning avatars from monocular videos, and can be applied in everyday activities for which setting a studio with multi-view cameras to capture training data would be highly challenging.
\begin{figure}[h]	
	\begin{center}
		\includegraphics[width=\linewidth]{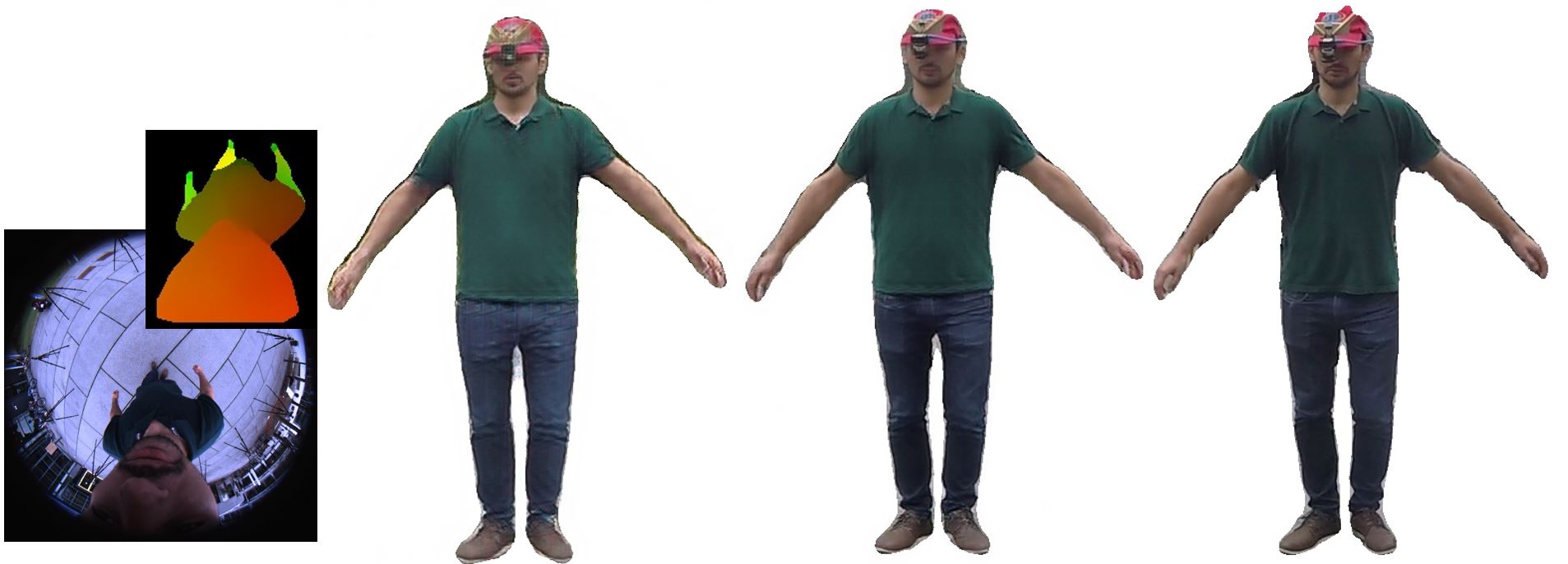}
	\end{center}
	\vspace{-0.18in}
	\caption{Renderings between single- and multi-video training.  From left to right: egocentric image and estimated DensePose, single-video result, multi-video result, ground truth. With multi-video training, more details are restored, such as the right hand.}
	\label{fig:mv_single}
	\vspace{-0.06in}
\end{figure}

\begin{table}[ht]
	\begin{tabular}{|c|c|c|c|c|c|c|}
		\hline
		\multirow{2}{*}{} & \multicolumn{3}{c|}{H1} & \multicolumn{3}{c|}{H4} \\ \cline{2-7} 
		& \small{SSIM}   & \small{LPIPS}  & \small{PSNR} & \small{SSIM}    & \small{LPIPS} & \small{PSNR} \\ \hline
		ImT                   & {8.134}  & \textbf{1.174} & 2.104 & \textbf{7.893} & \textbf{1.246} & \textbf{1.780} \\ \hline
		P2P                & 8.038 & 1.218 & 2.077 & 7.823  & 1.301 & 1.753 \\ \hline
		ExT                   & 7.993  & 1.258 & 2.059 & 7.716  & 1.374 & 1.722 \\ \hline
		Ego                 & 8.038   & 1.228 & 2.075 & 7.825  & 1.287 & 1.763 \\ \hline
		MV                  & \textbf{8.174}   & 1.217 & \textbf{2.140} & 7.770  & 1.343 & 1.732 \\ \hline
		FNet                  & 8.022  & 1.220 & {2.095} & 7.744  & 1.372 & 1.736 \\ \hline
	\end{tabular}
	\vspace{-0.1in}
	\caption{Quantitative comparisons of multi-video training on H1 outdoor and H4 indoor datasets. SSIM and LPIPS are multiplied by 10, PSNR by 0.1. The methods are listed as follows: Im-Tex, Pix2PixHD, Ex-Tex, Only-Ego, Only-MV, Fea-Net.  }
	\label{tab:quant_mv}
	\vspace{-0.12in}
\end{table}

\begin{figure}[ht]
	\begin{center}
		\includegraphics[width=\linewidth]{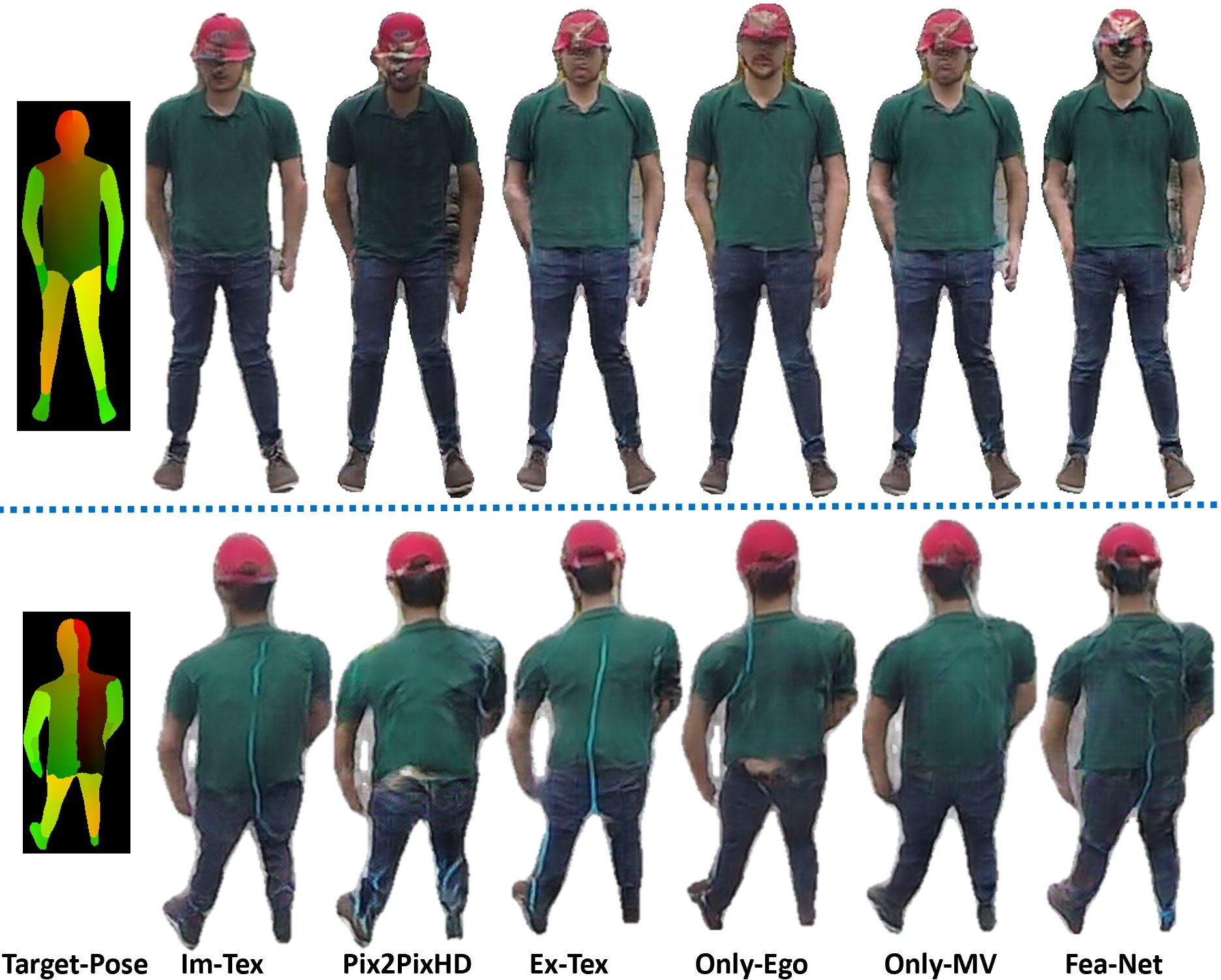}
	\end{center}
	\vspace{-0.18in}
	\caption{Renderings of different views in applications.}
	\label{fig:demo_cmp}
	\vspace{-0.15in}
\end{figure}

\noindent \textbf{Multi-video comparisons.} We also conduct multi-video experiments where each method was trained on multiple videos from different viewpoints in Table~\ref{tab:quant_mv}. We train the 6 methods with 9 multi-view cameras for H1, and 4 multi-view cameras for H4. Each method performs better when more video sequences are added in training, and especially our method can reconstruct more details, as shown in Figure \ref{fig:mv_single}. More comparisons on multi-video training are provided in Figure \ref{fig:demo_cmp} and the supplementary material. 

\subsection{Comparisons to Textured Neural Avatar (TNA)}
%
We compare our method against static texture based method TNA in Figure \ref{fig:comp_tna}, where ours can produce time- and pose-dependent appearance details, whereas TNA cannot. 

\begin{figure}	
	\begin{center}
		\includegraphics[width=\linewidth]{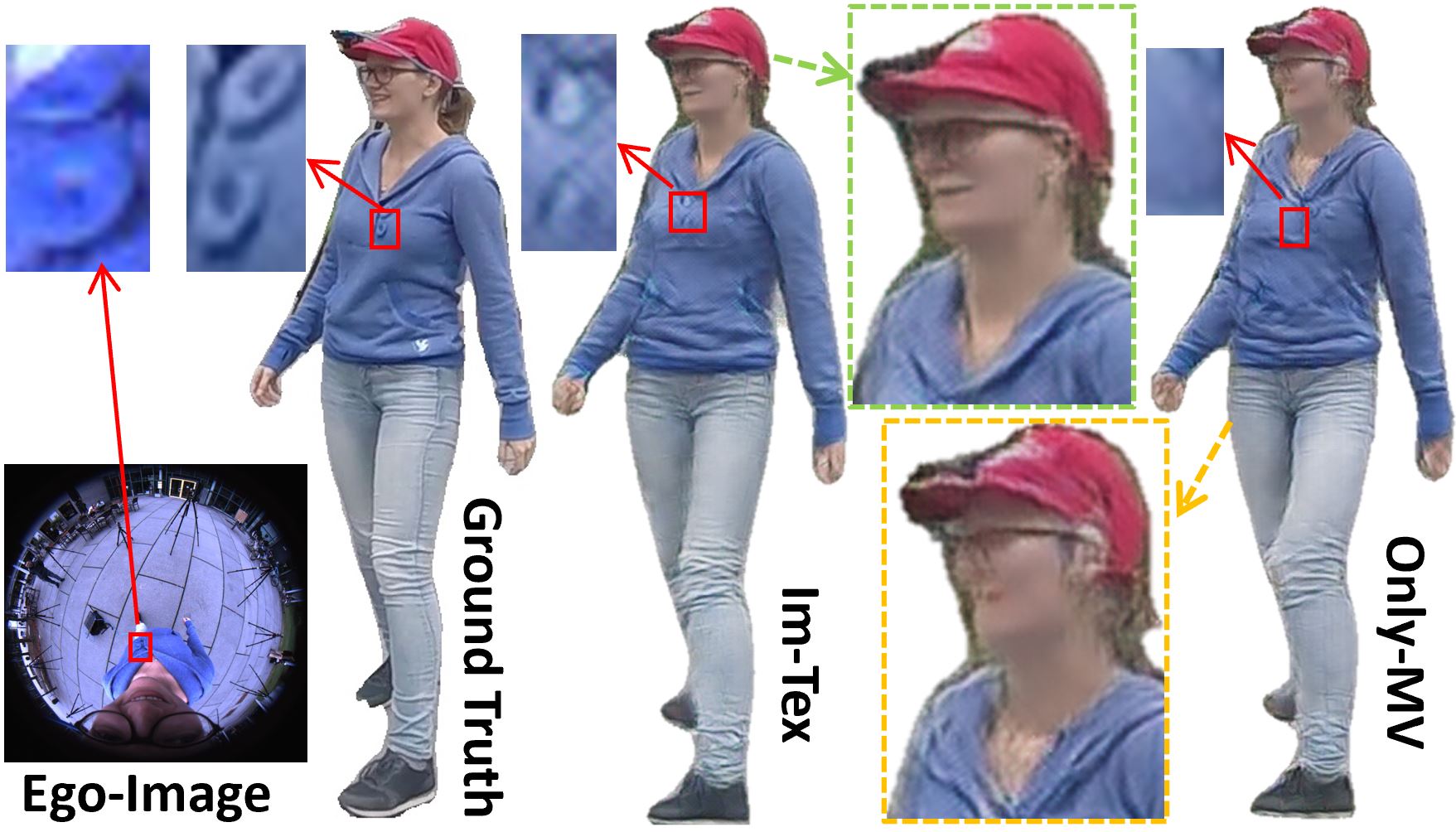}
	\end{center}
	\vspace{-0.2in}
	\caption{Compared with Only-MV, the use of egocentric texture of our Im-Tex leads to improvements visible in faces, neck and collar. We even restore the tiny clothes buttons visible in the Ego-Image.}
	\label{fig:comp_mv}
	\vspace{-0.02in}
\end{figure}

\begin{figure}	
	\begin{center}
		\includegraphics[width=\linewidth]{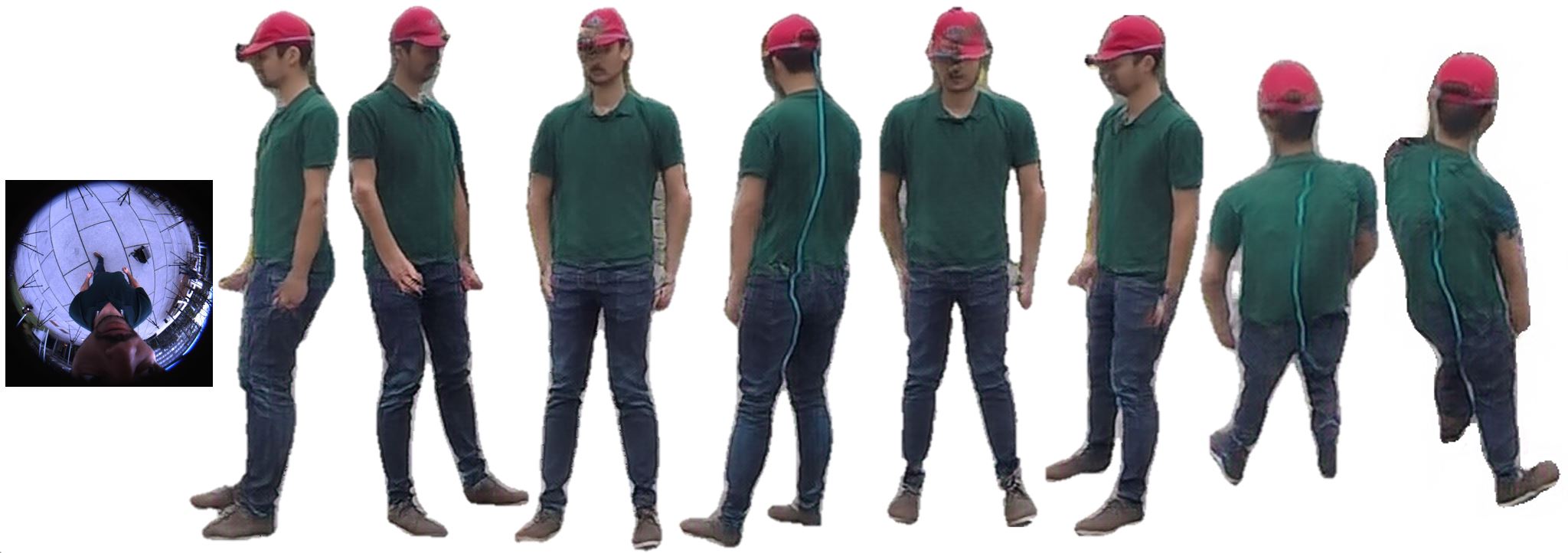}
	\end{center}
	\vspace{-0.2in}
	\caption{In applications, given an egocentric camera image (left), our methods can generate plausible avatars from different viewpoints, even with high elevation degree (the last four) unseen in training. We show the same example frame as Figure \ref{fig:overview}.}
	\label{fig:demo_all}
	\vspace{-0.1in}
\end{figure}

\begin{figure}	
	\begin{center}
		\includegraphics[width=\linewidth]{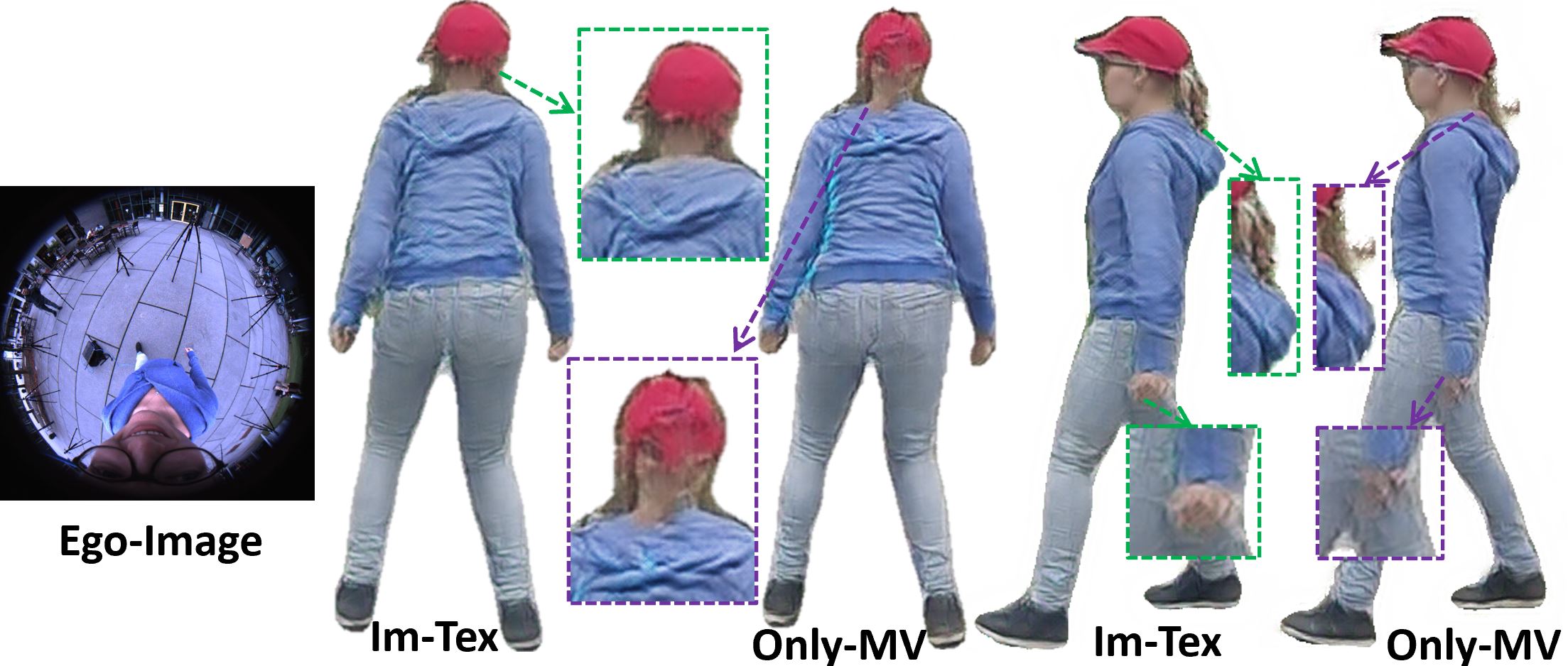}
	\end{center}
	\vspace{-0.18in}
	\caption{Comparisons with Only-MV on occluded parts.}
	\label{fig:comp_occluded}
	\vspace{-0.06in}
\end{figure}

%
%


\subsection{Ablation Study}
We study the advantages of the learned neural texture over other human rendering methods introduced in Section \ref{sec:baselines}, including Pix2PixHD \cite{pix2pixhd}, Fea-Net \cite{feanet}, Ex-Tex, Only-Ego, and Only-MV. Note that the main difference among these methods is the texture synthesis module. We get the following conclusions by comparing different methods according to the quantitative and qualitative results in Table \ref{tab:quant_single}, Figure \ref{fig:comp_method}-\ref{fig:demo_cmp} and our supplementary material. We use $a > b$ to denote that method $a$ outperforms method $b$. 

1) Texture synthesis methods (e.g. Im-Tex, Only-MV) outperform the direct pose-to-image translation method (Pix2PixHD), implied by \textbf{Only-MV $>$ Pix2PixHD}, where the only difference is that Only-MV has an implicit texture stack $T_m$.
2) Textures from egocentric images help improve the rendering quality, implied by \textbf{Im-Tex $>$ Only-MV}.
3) The implicit texture stack $T_m$ matters, implied by \textbf{Im-Tex $>$ Only-Ego}.
4) Implicit texture stacks capture detailed appearance variation
across poses and viewpoints better than explicit color texture stacks, implied by \textbf{Im-Tex $>$ Ex-Tex}.
5) For our human specific task, maintaining a globally shared texture stack works better than extracting temporary features from egocentric images for each frame, implied by \textbf{Im-Tex $>$ Fea-Net} \cite{feanet}. Different from Im-Tex, Fea-Net has a special FeatureNet to extract high dimensional features from egocentric inputs. Though Fea-Net has almost twice as many parameters as ours, its performance is not as good as ours on this person specific task.

\noindent \textbf{Why does the proposed method outperform
the others?} First,
we attribute this improvement to the explicit texture synthesis.
Compared with Pix2PixHD,
which learns implicit correlations between the target poses
and RGB avatars, our method explicitly utilizes the semantics
of the target poses by Feature Rendering and formulates
the rendering as an easier transformation from implicit textures
to avatars. Second, ours leveraging the Ego-Texture
as input outperforms Only-MV, as shown in Figure \ref{fig:comp_method}, \ref{fig:comp_mv}, where ours shows better realism (e.g. faces and clothes) and even reconstructs the tiny buttons visible in the egocentric images in Figure \ref{fig:comp_mv}. We also see that ours outperforms Only-MV even on occluded parts (e.g. the human head in row 1 of Figure \ref{fig:comp_method}, the back of human in the second row of Figure \ref{fig:demo_cmp}, and Figure \ref{fig:comp_occluded}), where the RGB textures
from the Ego-Texture are not directly used. However, the additional supervision contained in the Ego-Texture makes Render-Net learn better correlation between the occluded part and the visible texture regions, and possibly helps learn implicit shape information useful for more accurate renderings.

\noindent \textbf{Discussion: egocentric setup.} Why not first transform the fisheye images to normal images, then perform traditional human synthesis? Since we are using a fisheye camera with a $180^{\circ}$ field of view (FOV), it is not possible to project (in other words `undistort') our image to a rectilinear image. Although the cropped fisheye image (by reducing the FOV) can be undistorted, we lose the main advantage of using the fisheye image by doing so -- full body capturing under a wide range of motions, including fully extended arms, etc. This is also the reason many existing egocentric methods process the fisheye images directly \cite{mo2cap2, DBLP:conf/iccv/TomePAB19}.




\subsection{Applications}

In real applications, as introduced in Figure \ref{fig:overview} and Section \ref{sec:pose}, EgoRenderer uses Mo2Cap2 \cite{mo2cap2} to extract 3D poses and synthesize target poses. The synthesized target poses and comparisons of the six methods are provided in Figure \ref{fig:demo_cmp} for the example frame in Figure \ref{fig:overview}. Our method generates higher quality avatars than the other methods. In Figure \ref{fig:demo_all}, we show more viewpoints. It can be observed that EgoRenderer synthesizes reasonable avatars from novel viewpoints even with high elevation degrees. In addition, EgoRenderer can work in both local (human-centered) and global coordinate systems with camera tracking. See more details in the supplementary material.

\section{Summary and Discussion}
We proposed the EgoRenderer to synthesize free-viewpoint avatars from a single egocentric fisheye camera, and our system is an important step towards practical daily full-body reconstruction. We see our approach as the basis for many exciting applications in various areas, such as performance analysis, and human reconstructions in AR and VR. However, our system suffers from certain limitations mainly caused by the inaccuracies of Mo2Cap2 \cite{mo2cap2}: 1) Mo2Cap2 can estimate 15 joints (e.g., neck, shoulders, elbows, wrists, hips, knees, ankles and toes), whereas 24 joints are required to drive a SMPL model, which may cause inaccuracies in inverse kinematics and unnatural motions. 2) Mo2Cap2 per-frame predictions exhibit some {temporal instability} at run time, which causes temporal jittering in renderings. We believe these issues can be solved by optimizing the pose estimation module.

\noindent \textbf{Acknowledgements.} MZ was supported by NSF\#1813583. This work was partially funded by the ERC Consolidator Grant 4DRepLy (770784)

{\small
	\bibliographystyle{ieee_fullname}
	\bibliography{egbib}
}

\clearpage


\onecolumn
\appendix
\section*{Appendices}
\addcontentsline{toc}{section}{Appendices}
\renewcommand{\thesubsection}{\Alph{subsection}}
\section{Dataset}

\noindent \textbf{More details of the proposed synthetic dataset.} To render our synthetic dataset, we animated characters using the SMPL model \cite{smpl} with around 3000 different motions sampled from the CMU MoCap \cite{CMUmocap} dataset. More than 600 body textures were randomly chosen from the texture set provided by the SURREAL \cite{surreal} dataset. In total, we rendered 178,800 images for training.

\noindent \textbf{Dataset Pre-processing} After we captured datasets, we first synchronized the egocentric fisheye camera and multi-view cameras to register them in time, and used the pre-trained PointRend \cite{Kirillov2020PointRendIS} for foreground segmentation. 

\section{More Experimental Results}


\subsection{Quantitative Comparisons}


\noindent \textbf{Single-video comparisons.} We also provide the additional L1 distances (Table \ref{tab:quant_single_l1}) of each method on single-video datasets.

\begin{table}[ht]
	\centering
	\begin{tabular}{|c|c|c|c|c|c|c|}
		\hline
		\multirow{1}{*}{} & Im-Tex & Pix2PixHD \cite{pix2pixhd} & Ex-Tex & Only-Ego & Only-MV & Fea-Net \cite{feanet} \\ \hline
		H1 & {0.994} & {0.994} & \textbf{0.950} & 0.995 & 0.999 & 0.998 \\ \hline
		H2 & 1.191 & 1.199 & 1.215 & 1.233  & 1.224 & \textbf{1.175} \\ \hline
		H3 & \textbf{1.448} &1.522 & 1.562  & 1.484 & 1.516  & 1.511 \\ \hline 
		H4 & \textbf{0.894} & 0.906 & 0.944 & 0.911 & 0.905 & 0.926 \\ \hline
	\end{tabular}
	\caption{L1 distances of single-video training on different datasets. Numbers are multiplied by 10} 
	\label{tab:quant_single_l1}
\end{table}

\noindent \textbf{Multi-video comparisons.} The L1 distances of multi-video experiments for indoor (H1) and outdoor scenes (H4) are provided in Table~\ref{tab:quant_mv_l1}, where each method was trained on multiple videos from different viewpoints, 9 multi-view cameras for H1, and 4 multi-view cameras for H4.


\begin{table}[ht]
	\centering
	\begin{tabular}{|c|c|c|c|c|c|c|}
		\hline
		\multirow{1}{*}{} & Im-Tex & Pix2PixHD \cite{pix2pixhd} & Ex-Tex & Only-Ego & Only-MV & Fea-Net \cite{feanet} \\ \hline
		H1     & 0.645 & 0.650 & 0.688 & 0.651 & 0.646  & \textbf{0.641} \\ \hline
		H2 & \textbf{0.767} & 0.791 & 0.836 & 0.790 & 0.817 & 0.834 \\ \hline
	\end{tabular}
	\caption{L1 distances of multi-video training on H1 outdoor and H4 indoor datasets. Numbers are multiplied by 10} 
	\label{tab:quant_mv_l1}
\end{table}

\subsection{Local and Global Coordinate System}
\begin{figure*}[ht]
	\centering
	\begin{subfigure}[b]{0.9\textwidth}
		\centering
		\includegraphics[width=\textwidth]{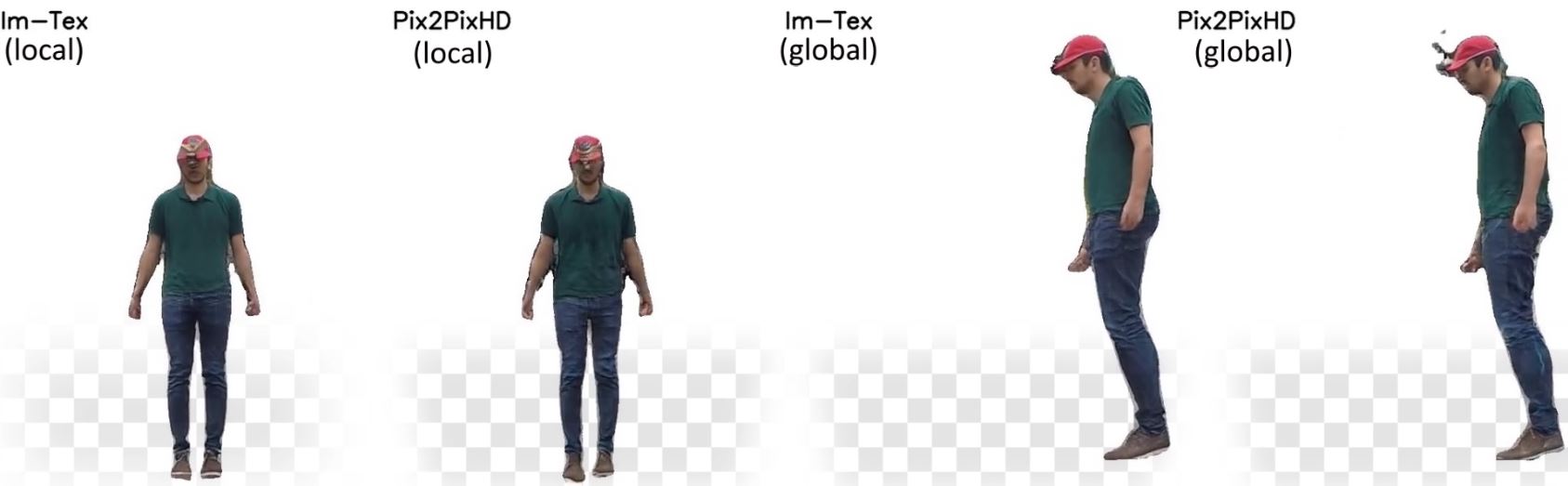}
		\caption{Renderings at timestamp 1.}
		\label{fig:y equals x}
	\end{subfigure}
	\hfill
	\begin{subfigure}[b]{0.9\textwidth}
		\centering
		\includegraphics[width=\textwidth]{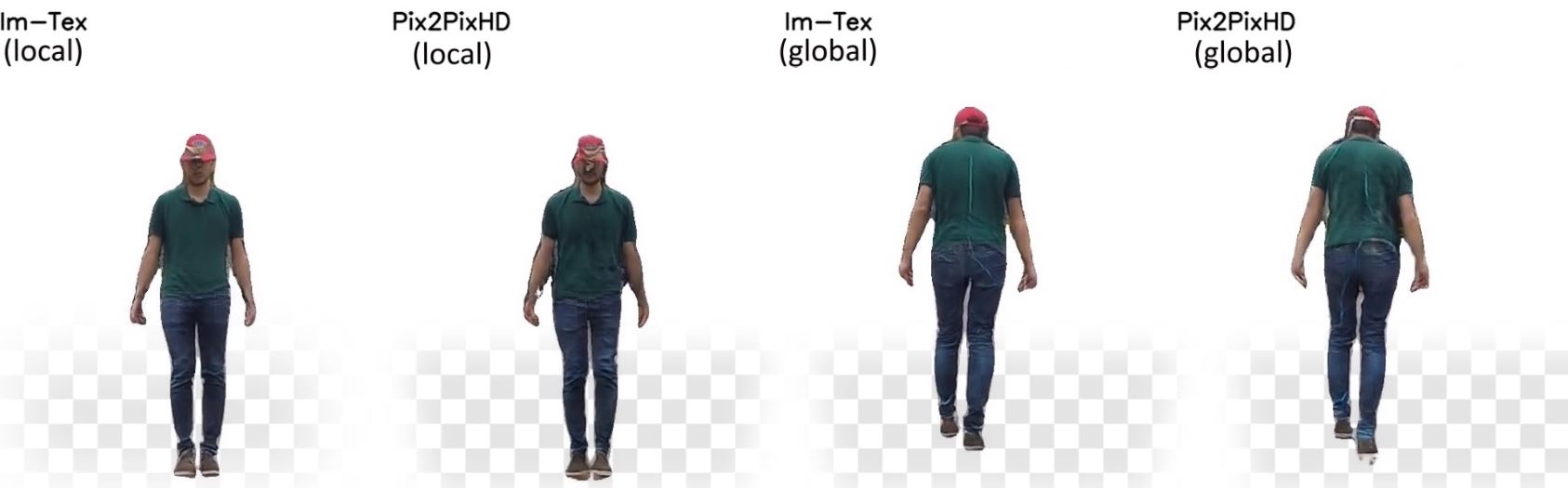}
		\caption{Renderings at another timestamp.}
		\label{fig:three sin x}
	\end{subfigure}
	\hfill
	\caption{EgoRenderer can work in both local (left) and global coordinate systems (right).}
	\label{fig:global_space}
\end{figure*}

Our system, EgoRenderer can work in both local (human-centered coordinate) and global coordinate systems (Figure \ref{fig:global_space}). For local system (left), we assume each user-specific viewpoint is relative
to the human. For global system (right), we synthesize global target poses by integrating local poses estimated by Mo2Cap2 and global tracking by external devices. More results can be found in the video demo.

\end{document}

%% file: intro_new.tex
The goal of this work is to render full-body avatars with realistic appearance and motion of a person wearing an egocentric fisheye camera from arbitrary external camera viewpoints (Figure \ref{fig:teaser}). Such egocentric capture and rendering enables new applications in sport performance analysis or health care.
Real-time free-viewpoint rendering of self-embodied avatars is also important in virtual reality (VR) and augmented reality (AR) applications, notably telepresence. 
A key advantage of our approach is that it uses a lightweight and compact sensor that could be mounted to glasses, headsets or caps, and that it is fully mobile. Therefore, actors can freely roam and are not limited to stay in confined spaces visible to external multi-camera setups.  


We approach free-viewpoint neural avatar rendering from our egocentric view by a combination of new solutions for egocentric pose estimation, appearance transfer, and free-viewpoint neural rendering; each of these need to be tailored to the challenging egocentric top-down fisheye perspective with strong distortions and self-occlusions.  
Most established pose estimation methods employ external outside-in camera views \cite{Mehta2017Monocular3H, Mehta2017VNectR3, Shotton2011RealtimeHP, pose, densepose} and are not directly applicable to our setting.  

\begin{figure}
	\begin{center}
		\includegraphics[width=0.86\linewidth]{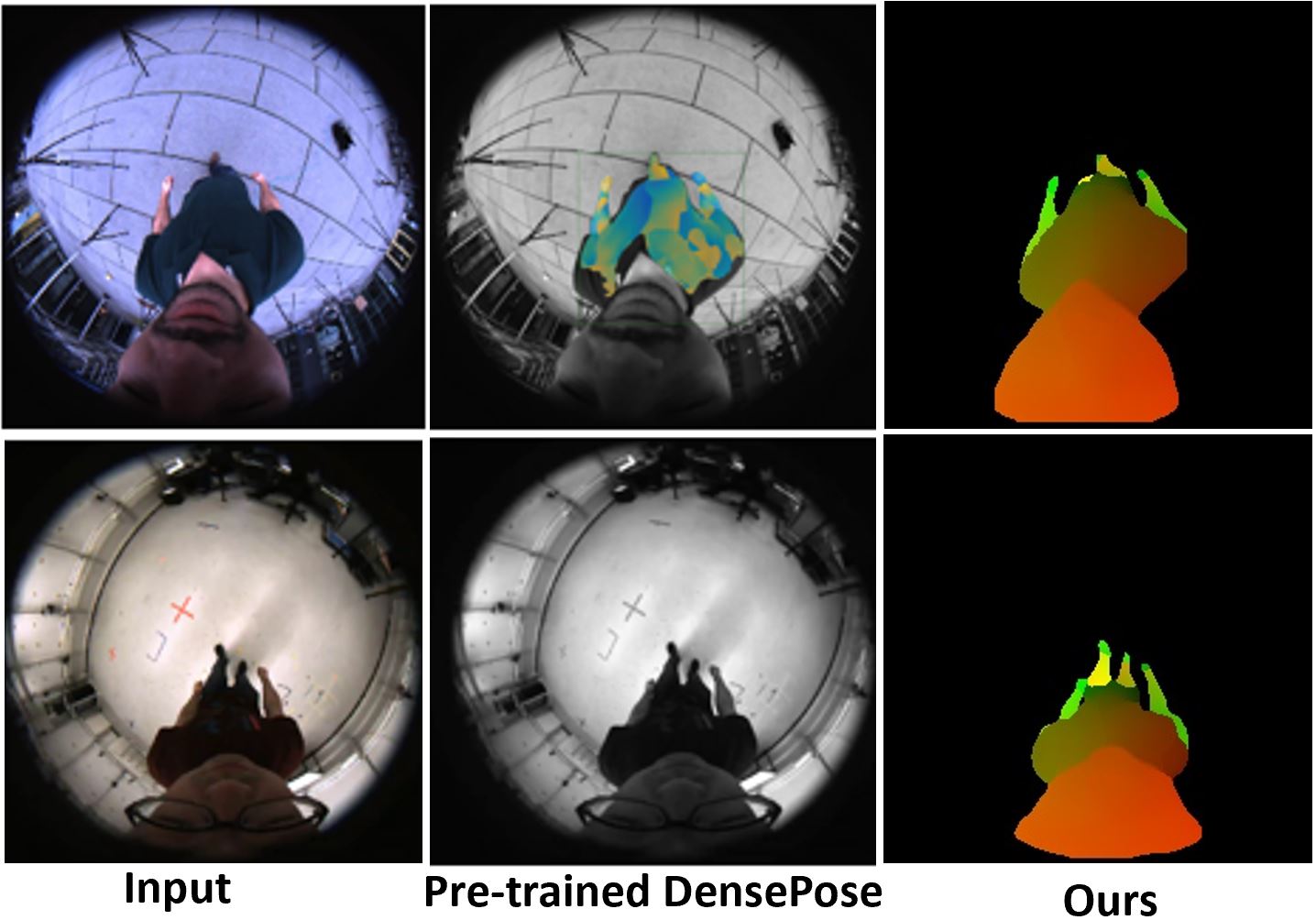}
	\end{center}
	\vspace{-0.17in}
	\caption{DensePose \cite{densepose} performs poorly on images captured by our setup, sometimes failing to  detect the human (second row). Our DensePose predictions are on the right.}
	\label{fig:comp_dp}
	\vspace{-0.15in}
\end{figure}
\begin{figure*}
	\begin{center}
		\includegraphics[width=\linewidth]{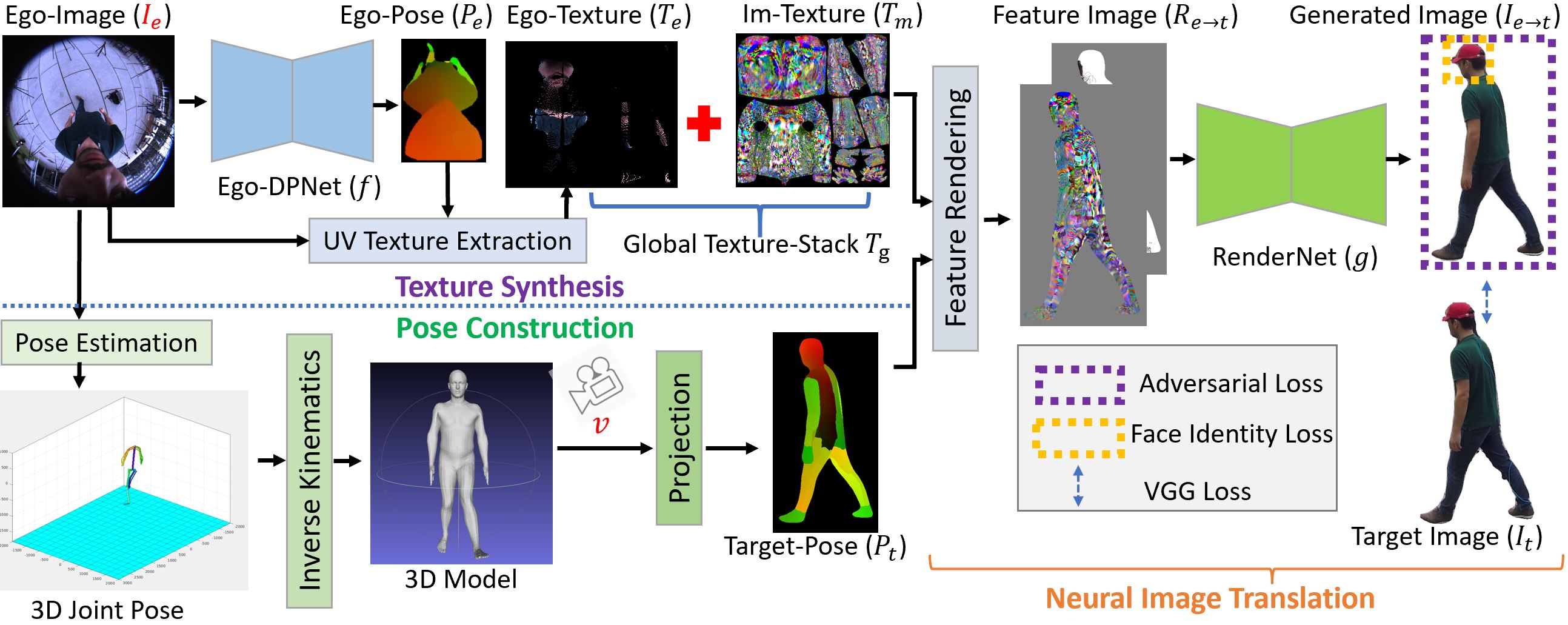}
	\end{center}
	\vspace{-0.13in}
	\caption{Pipeline overview. Given an egocentric image $I_e$ of a person and a user-defined viewpoint $v$, EgoRenderer synthesizes the full body avatar $I_{e \rightarrow t}$ from viewpoint $v$.
		EgoRenderer is custom tailored to our cap-mounted fisheye camera setup, and decomposes the rendering process into texture synthesis, pose construction and neural image translation. For texture synthesis, given an input image $I_e$ of a person in egocentric  camera space, we first predict the dense correspondences ($P_e$) between the input
		image $I_e$ and an underlying parametric mesh model using Ego-DPNet, which allows a partial UV Texture-map $T_e$ to be extracted for the body regions visible in the image. We also learn an implicit texture stack ($T_m$) during training, and concatenate $T_e$ and $T_m$ as the global texture representation $T_g$. For pose construction, given a user-defined viewpoint $v$, we synthesize a target pose image $P_t$ by projecting the parametric model from egocentric space to the target viewpoint. In the \textit{Feature Rendering} step, the parametric body mesh is textured with the global texture stack $T_g$ to produce intermediate feature images $R_{e \rightarrow t}$. An image-translation network RenderNet converts the  feature images to the final realistic image $I_{e \rightarrow t}$.}
	\label{fig:overview}
		\vspace{-0.1in}
\end{figure*}

Some recent approaches are designed to estimate 3D skeletal pose from head-mounted fisheye cameras~\cite{mo2cap2, Tom2019xREgoPoseE3}. However, our setting requires a denser pixel wise estimation of egocentric pose and shape, as dense correspondences are prerequisite to transfer the texture appearance of a person from egocentric to external views (Figure \ref{fig:comp_dp}).
Similarly, recent neural rendering-based pose transfer methods enable creation of highly realistic animation videos of humans under user-specified target motion~\cite{everybody,Aberman2019DeepVP,Liu2018NeuralAA,wang2018vid2vid}. However, all of these are tailored to external oustside-in views, such that target motions already need to be specified as skeletal pose or template mesh sequences from the extrinsic camera view. We face the additional challenge of transferring appearance and pose from the starkly distorted egocentric view to the external view. 
To enable highly realistic appearance and pose transfer of the actor wearing the camera to an arbitrary external view, even in more general scene conditions, EgoRenderer decomposes the rendering pipeline into \textbf{texture synthesis}, \textbf{pose construction}, and \textbf{neural image translation}, as shown in Figure~\ref{fig:overview}. 

\begin{figure}
	\begin{center}
		\includegraphics[width=0.9\linewidth]{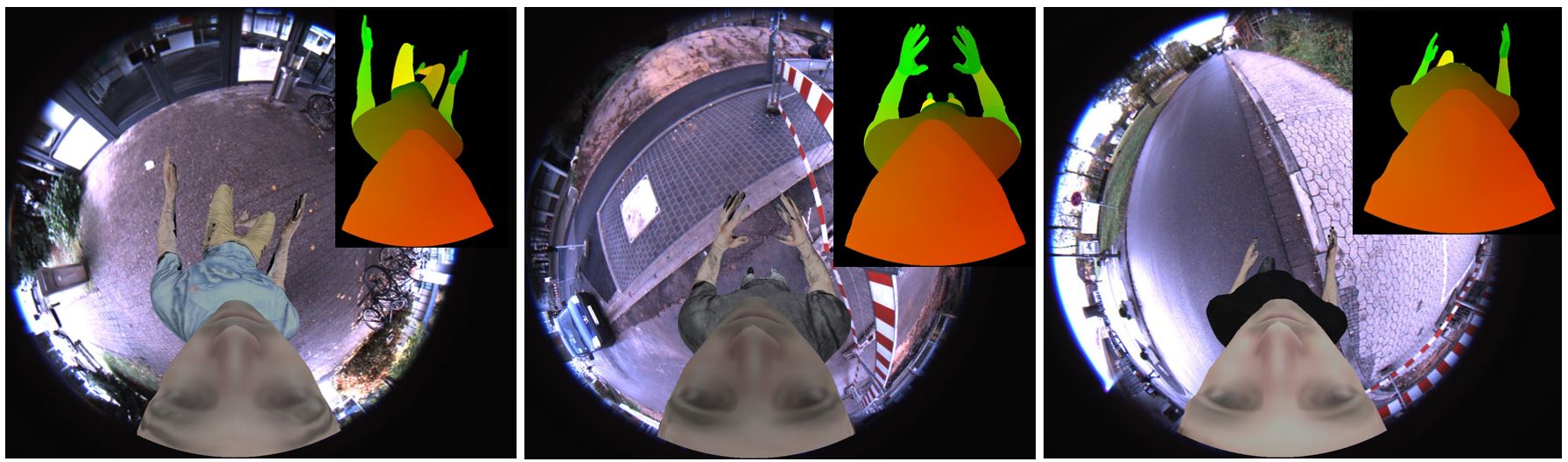}
	\end{center}
	\vspace{-0.17in}
	\caption{Examples from our synthetically rendered fisheye training dataset (top-right: ground truth DensePose). Our dataset features a large variety of poses, human body appearance, and realistic backgrounds.}
	\label{fig:synthetic}
	\vspace{-0.02in}
\end{figure}

\textbf{Texture Synthesis}. In contrast to most aforementioned image-based pose transfer methods for outside-in views~\cite{everybody,texava}, EgoRenderer explicitly builds an estimation of surface textures of a person on top of a parametric body model. 
Specifically, we extract explicit (color) textures from egocentric images and learn implicit textures from a multi-view dataset in a training phase. We then combine them to form our full texture representation of the person. Compared to static color texture maps, the learned implicit textures better capture detailed appearance variation across poses and viewpoints. To extract (partial) textures for visible body parts from egocentric images, we create a large synthetic dataset (see Figure \ref{fig:synthetic}) and train an Ego-DPNet network tailored to our setup to infer the dense correspondences between the input egocentric images and an underlying parametric body model, as shown in Figure \ref{fig:comp_dp}.

\textbf{Pose Construction}. Different from earlier neural human rendering methods that expect target poses as input, irrespective of where these targets are from~\cite{texava}, EgoRenderer works end-to-end. We have to exactly reproduce pose and appearance seen in an egocentric image from any external viewpoint. We support neural rendering of the target view by projecting the 3D parametric model from egocentric camera space to the target viewpoint, enabling us to also transfer the  
partially visible texture appearance. 

\textbf{Neural Image Translation}. Pose construction enables us to render the 3D model in the desired external view using both implicit textures and color textures. We transform these images into the final color image by means of a neural image translation network. 
Experiments show that this implicit to explicit rendering approach outperforms direct pose-to-image translation.

Our qualitative and quantitative evaluations demonstrate that our EgoRenderer system generalizes better to novel viewpoints and poses than baseline methods on our test set. To summarize, our contributions are as follows:

\noindent 1) A large synthetic ground truth training dataset of top-down fisheye images and an Ego-DPNet network tailored to our fisheye camera setup to predict dense correspondence to a parametric body model from egocentric images.

\noindent 2) An end-to-end EgoRenderer system that takes single egocentric images as input and generates full-body avatar renderings from external user-defined viewpoints.

%% file: related_work.tex
\section{Related Work}
Our approach is closely related to many sub-fields of visual computing, and below we discuss a small subset of these connections.

\noindent \textbf{Neural Rendering.} Neural rendering is a class of deep image and video generation approaches that combines generative machine learning techniques with physical knowledge from computer graphics to obtain controllable outputs. Many neural rendering methods \cite{sitzmann2019deepvoxels, Sitzmann2019, mildenhall2020nerf, Liu2020NeuralSV, Thies2019} learn implicit representations of scene instead of explicitly modeling geometry, such as DeepVoxels \cite{sitzmann2019deepvoxels}, SRNs \cite{Sitzmann2019}, NeRF \cite{mildenhall2020nerf} and NSVF \cite{Liu2020NeuralSV}.
However, only a few neural scene representations handle dynamic scenes ~\cite{Kim2018,lombardi2019neural}. Our neural rendering approach is inspired by the aforementioned scene specific methods. To handle the dynamic nature of the scene, we learn a person specific implicit texture map on a parametric model of humans. This learned implicit texture stack, along with the pose dependent appearance from the egocentric camera, is used by a high fidelity generator to produce realistic renderings of the person.

\noindent \textbf{Pose Transfer and Human Re-enactment.} Pose transfer, first introduced by~\cite{MaSJSTV2017}, refers to the problem of re-rendering a person from a different viewpoint and pose from the appearance of a single image. Most approaches formulate this problem as an image-to-image mapping problem, i.e. given a reference image of a target person, mapping the body pose in the form of renderings of a skeleton~\cite{everybody,SiaroSLS2017,Pumarola_2018_CVPR,KratzHPV2017,zhu2019progressive}, dense mesh~\cite{Liu2019,wang2018vid2vid,liu2020NeuralHumanRendering,feanet,Neverova2018,Grigorev2019CoordinateBasedTI} or joint position heatmaps~\cite{MaSJSTV2017,Aberman2019DeepVP,Ma18}, to real images. 
To better map the appearance of the reference to the generated image, some methods~\cite{liu2020NeuralHumanRendering,feanet} first map the person appearance in screen space to UV space and feed the rendering of the person in the target pose with the UV texture map into an image translation network. However, these methods generally works in external camera setup with regular viewpoints. We propose a new method that extracts appearance from the top-down egocentric fisheye camera using a novel network, and combines it with a learned person specific neural texture for a high fidelity generation. 

\noindent \textbf{Egocentric Systems.} Because of their mobility and flexibility, egocentric systems have made significant progress in recent years. Applications of egocentric systems can be divided into face, gesture, and full body. 
\cite{Elgharib2020egocentricconferencing,elgharib2019egoface,thies16FaceVR, Lombardi2018DeepFace, cha2018fullycapture, DBLP:conf/uist/SuganoB15, DBLP:journals/tog/LiTOWTHNM15} study face estimation with a head-mounted camera. \cite{Singh2016ActionRecon,SridharFastHandTracker,DBLP:conf/cvpr/MaFK16,DBLP:conf/iccv/CaoZ0LC17,OccludedHands_ICCV2017,DBLP:conf/cvpr/OhnishiKKH16, DBLP:journals/pr/SinghAJ17} perform gesture and activity recognition by using a head-mounted or chest-worn camera. In the case of full body, most of the methods either opt for an inside-out configuration \cite{DBLP:journals/tog/ShiratoriPSSH11,DBLP:conf/cvpr/JiangG17a,yuan20183d,DBLP:conf/iccv/YuanK19}, or employ fisheye cameras for a large field of view \cite{mo2cap2,DBLP:journals/tog/RhodinRCISSST16}. Rhodin et al. \cite{DBLP:journals/tog/RhodinRCISSST16} propose the first full-body capture method with a helmet-mounted stereo fisheye camera pair. Xu et al. \cite{mo2cap2} and Tome et al. \cite{Tom2019xREgoPoseE3} use a much more compact and flexible monocular setting for full body pose estimation. However, our setting learns a denser pixel wise estimation of egocentric pose and shape (see Figure \ref{fig:comp_dp}).
